\documentclass[a4paper,fleqn]{cas-dc}

\usepackage[numbers]{natbib}

\def\tsc#1{\csdef{#1}{\textsc{\lowercase{#1}}\xspace}}
\tsc{WGM}
\tsc{QE}
\tsc{EP}
\tsc{PMS}
\tsc{BEC}
\tsc{DE}

\usepackage{color}
\usepackage{xcolor}

\begin{document}
\let\WriteBookmarks\relax
\def\floatpagepagefraction{1}
\def\textpagefraction{.001}
\shorttitle{Generative Object Insertion using Multi-view Diffusion}
\shortauthors{H. Zhong et~al.}

\title[mode = title]{Generative Object Insertion in Gaussian Splatting with a Multi-View Diffusion Model}


\author[1]{Hongliang Zhong}[type=editor,
                        orcid=0009-0002-0840-8812]
\ead{hlzhong2-c@my.cityu.edu.hk}

\author[1]{Can Wang}[type=editor,
                        orcid=0000-0002-5102-1464]
\ead{cwang355-c@my.cityu.edu.hk}

\author[1]{Jingbo Zhang}[type=editor,
                        orcid=0000-0003-0009-2315]
\ead{jbzhang6-c@my.cityu.edu.hk}

\author[1]{Jing Liao}[type=editor,
                        orcid=0000-0001-7014-5377]
\cormark[1]
\ead{jingliao@cityu.edu.hk}

\credit{Conceptualization of this study, Methodology, Software}

\affiliation[1]{organization={Department of Computer Science, City University of Hong Kong},
                city={Hong Kong},
                country={China}}

\cortext[cor1]{Corresponding author}

\begin{abstract}
Generating and inserting new objects into 3D content is a compelling approach for achieving versatile scene recreation. Existing methods, which rely on SDS optimization or single-view inpainting, often struggle to produce high-quality results. To address this, we propose a novel method for object insertion in 3D content represented by Gaussian Splatting. Our approach introduces a multi-view diffusion model, dubbed MVInpainter, which is built upon a pre-trained stable video diffusion model to facilitate view-consistent object inpainting. Within MVInpainter, we incorporate a ControlNet-based conditional injection module to enable controlled and more predictable multi-view generation. After generating the multi-view inpainted results, we further propose a mask-aware 3D reconstruction technique to refine Gaussian Splatting reconstruction from these sparse inpainted views. By leveraging these fabricate techniques, our approach yields diverse results, ensures view-consistent and harmonious insertions, and produces better object quality. Extensive experiments demonstrate that our approach outperforms existing methods.
\end{abstract}


\maketitle

\section{Introduction}

Recent advances in novel view synthesis, particularly in techniques such as neural radiance fields (NeRFs) \cite{mildenhall2021nerf,barron2021mip,barron2022mip,wang2021nerf,Martin-Brualla_2021_CVPR} and Gaussian Splatting \cite{kerbl20233d,qin2024langsplat,yan2024multi,wu20244d}, have significantly propelled the progress in 3D reconstruction applications \cite{long2024wonder3d,wang2022mdisn,butime20063d}. Consequently, the demand for creating new objects in a 3D content to achieve versatile re-creation is rising rapidly. These advancements have opened new possibilities for enhancing the fidelity and usability of reconstructed scenes, benefiting virtual reality, gaming, and digital content creation. However, generating and inserting new objects into a 3D scene remains a challenging task due to the following reasons: 
1) Ensuring 3D-consistent generation and placement of objects from different viewpoints.
2) Producing high-quality new objects with fabricate geometry and texture
3) Achieving harmony between the inserted object and the existing scene.

Motivated by the success of diffusion models \cite{rombach2022high,yuan2024diffmat,brooks2023instructpix2pix,kawar2023imagic,zhang2023adding,zhao2024uni,chen2023controlstyle}, various approaches \cite{dai2024go,chen2024mvip,bartrum2024replaceanything3d} have sought to tackle theses challenges using Score Distillation Sampling (SDS) \cite{poole2022dreamfusion} of diffusion models. The core principle involves optimizing 3D representations by encouraging their rendered images to move toward high-probability density regions conditioned on text, with supervision provided by a pre-trained 2D diffusion model \cite{yu2023text}. Building on this principle, new objects can be created by optimizing the 3D representation in alignment with the direction guided by the SDS process.
These methods can produce view-consistent and harmonious 3D objects from diverse text prompts without requiring 3D data for training. However, SDS optimization often suffers from high optimization randomness and saturation issues \cite{liu2024humangaussian,chen2024gaussianeditor}, leading to less satisfactory visual quality. In contrast, our method employs a multi-view diffusion model to inpaint view-consistent objects for 3D generation, thereby avoiding the issues associated with SDS optimization and achieving high-quality, realistic results.

There are approaches that use image inpainting and 3D reconstruction to insert new objects into a 3D scene without using SDS optimization. 
Some methods \cite{liu2024infusion,mirzaei2023reference} inpaint the target object on a single view, then use the estimated depth to warp this object into 3D space, serving as an initialization \cite{liu2024infusion} or supervised signals \cite{mirzaei2023reference} for 3D object reconstruction. After inpainting the target object, more intuitive methods \cite{shahbazi2024inserf,chen2024gaussianeditor} utilize a single-view 3D reconstruction model to obtain the target 3D object from its 2D appearance. This generated 3D object is then placed into the original 3D scene according to the depth information. These methods excel at creating realistic and consistent objects but focus solely on harmonization within the inpainting view, failing to achieve consistent results across all viewpoints. In contrast, our multi-view diffusion model ensures harmonious inpainting across multiple viewpoints.

In summary, we propose a multi-view diffusion model for generative object insertion. Specifically, given a pre-trained 3D scene representation (we use Gaussian Splatting for its ability in fast and high-quality novel view synthesis), a 3D bounding box (BBox) indicating the target location, and a textual description of the target object, we first use SDS to obtain a coarse model, similar to other SDS-based methods \cite{poole2022dreamfusion,dai2024go}.
Next, we derive backgrounds, BBox-level masks, and depth maps from both the original 3D scene and the coarse model. We then design a multi-view diffusion model, MVInpainter, that takes these three sets along with the text prompt as input to produce view-consistent inpainting results that match the target object description. Finally, we propose a mask-aware reconstruction method that effectively leverages both the generated multi-view inpainted results and the original views to reconstruct 3D Gaussian Splatting. This approach mitigates the floating and noisy artifacts typically resulting from sparse view reconstruction, thereby enabling the rendering of higher-quality novel viewpoints.

We evaluate the proposed method on various complex scenes and multiple datasets \cite{barron2022mip,mirzaei2023spin}. Our experimental results demonstrate the effectiveness of our method in inserting diverse, view-consistent, harmonious, and high-quality objects into complex 3D scenes. To summarize, our key contributions are:

\begin{itemize}
\item We introduce the first multi-view diffusion model-based 3D object insertion method that sets a new standard in the field. 

\item We present a novel multi-view diffusion model that fully leverages background information, BBox masks, and depth to generate high-quality, view-consistent, and background-harmonious object insertions. Additionally, we propose a mask-aware 3D reconstruction method to enhance Gaussian Splatting reconstruction performance from sparse inpainted views, leading to better rendering quality.

\item Our experiments and visualizations demonstrate the advantages of the proposed method for generative object insertion compared to existing approaches. Specifically, our approach yields diverse results, ensures view-consistent and harmonious insertions, and produces better object quality.
\end{itemize}

\section{Related Work}

\subsection{3D Content Editing}
 
3D content editing \cite{zhuang2023dreameditor,chen2021towards,sun2024nerfeditor,lu2024advances} has experienced significant advancements, particularly in the modification  \cite{xu2024texture,zhao2022metaverse} of geometry and appearance based on textual and spatial guidance. Several recent methods have shown remarkable promise using various inputs such as text prompts \cite{park2023ed}, sketches \cite{mikaeili2023sked}, and reference images \cite{bao2023sine,dong2024vica,fujiwara2024style}, as well as leveraging models like CLIP \cite{wang2022clip,wang2023nerf,hyung2023local} and diffusion models \cite{park2023ed,haque2023instruct,rombach2022high,yuan2024diffmat,brooks2023instructpix2pix,kawar2023imagic,zhang2023adding,zhao2024uni,chen2023controlstyle,xie2023smartbrush,yang2023uni} to provide editing directions. Among these works, diffusion-based methods, in particular, have significantly enhanced the effectiveness of 3D content editing by utilizing the strong capabilities of diffusion models in text-based generation and editing.
To enhance the controllability of 3D editing, SketchDream \cite{liu2024sketchdream} introduces a sketch-based multi-view image generation diffusion model, enabling sketch-based text-to-3D generation and editing in object-centric cases. On the other hand, QNeRF \cite{patashnik2024consolidating} suggests extracting query features of the self-attention layers in the diffusion model during multi-view editing, then training a neural field to consolidate these queries, thereby enhancing editing consistency. However, these methods either primarily focus on localized edits, altering existing objects in the 3D scene, or solely support scene-independent, object-centric 3D content generation. This poses challenges for creating new objects from scratch or removing objects within a given scene.
In recent years, there has been a growing body of work focusing on object removal in 3D scenes by exploiting techniques such as differentiable mesh optimization \cite{wang2023mesh}, multi-view segmentation guidance \cite{yin2023or,mirzaei2023spin}, and depth priors \cite{liu2022nerf}. These methods typically involve delineating the editing boundaries in different viewpoints based on the shape of the target object. Consequently, they are unable to generate new objects in a scene from scratch without new object-related priors \cite{shahbazi2024inserf}.
In contrast, our work targets the creation of 3D-consistent, high-quality, and background-harmonious new objects in 3D scenes from scratch.

\subsection{SDS-based Object Insertion}

To facilitate object generation and insertion in 3D scenes, some studies \cite{dai2024go, chen2024mvip, bartrum2024replaceanything3d} have integrated the classic Score Distillation Sampling (SDS) algorithm \cite{poole2022dreamfusion} into 3D generation. This integration aims to leverage SDS's robust generative capabilities to directly produce viewpoint-consistent 3D content within scenes. Drawing upon prior knowledge from 2D diffusion, SDS optimizes the rendered images of 3D models to align with the desired image distribution conditioned on text. This ensures that the edited 3D model meets the desired outcomes across various viewpoints, enabling the generation and insertion of 3D content from scratch.
ReplaceAnything3D \cite{bartrum2024replaceanything3d} enhances this approach by incorporating an improved version of SDS, known as HiFA \cite{zhu2023hifa}, for generative object insertion, thereby enhancing the realism of editing effects. 
TIP-Editor \cite{zhuang2024tip} introduces a stepwise 2D personalization strategy and a 3D scene editing framework incorporating SDS loss. This system enables 3D editing, such as object insertion, guided not only by the text prompt but also by a reference image.
MVIP-NeRF \cite{chen2024mvip} proposes a multi-view SDS that concurrently supervises surface normals and scene appearance, providing stronger supervision for scene editing and achieving improved results.
While SDS-based methods effectively achieve viewpoint-consistent 3D content generation and editing, the inherent stochasticity in their optimization process often results in noticeable quality issues, such as over-smoothness \cite{liang2024luciddreamer,wang2024prolificdreamer} and high saturation levels \cite{liu2024humangaussian, chen2024gaussianeditor}. In contrast, our method avoids these issues by using SDS optimization solely to provide coarse initial priors rather than the final editing output. We then design a multi-view diffusion model to inpaint view-consistent editing results.

\subsection{Combine Image Inpainting
and 3D Reconstruction for Object Insertion}

In the quest for high-quality 3D content generation and insertion, several methods \cite{rojas2024datenerf, mirzaei2023reference, chen2024gaussianeditor, liu2024infusion,shahbazi2024inserf} have explored 
combining image inpainting and 3D reconstruction for 3D object insertion. For instance, InFusion \cite{liu2024infusion} starts by inpainting and inserting the object in 2D from a single reference view. It then projects this inpainted 2D appearance back into the 3D space of the original scene and performs a brief reconstruction optimization in the reference viewpoint to produce the final edited scene. On the other hand, after inpainting the object in an image, GaussianEditor \cite{chen2024gaussianeditor} first reconstructs the complete shape of the target object from its appearance in this reference image using a single-view 3D reconstruction model \cite{long2024wonder3d}. It then predicts the depth and integrates the object back into the original scene. Both approaches are limited by their reliance on single-viewpoint inpainting, which can undermine the seamless integration of the reconstructed object with the environment from other viewpoints. To tackle these challenges, we present the MVInpainter model along with a mask-aware 3D reconstruction method, specifically designed to produce 3D-consistent, high-quality, and harmonious object insertions across multiple viewpoints. Building on these innovative techniques, our approach ensures a more seamless and detailed integration of objects into the scene.

\section{Generative Object Insertion}\label{insertion}

\begin{figure*}
    \centering
    \includegraphics[width=.95\textwidth]{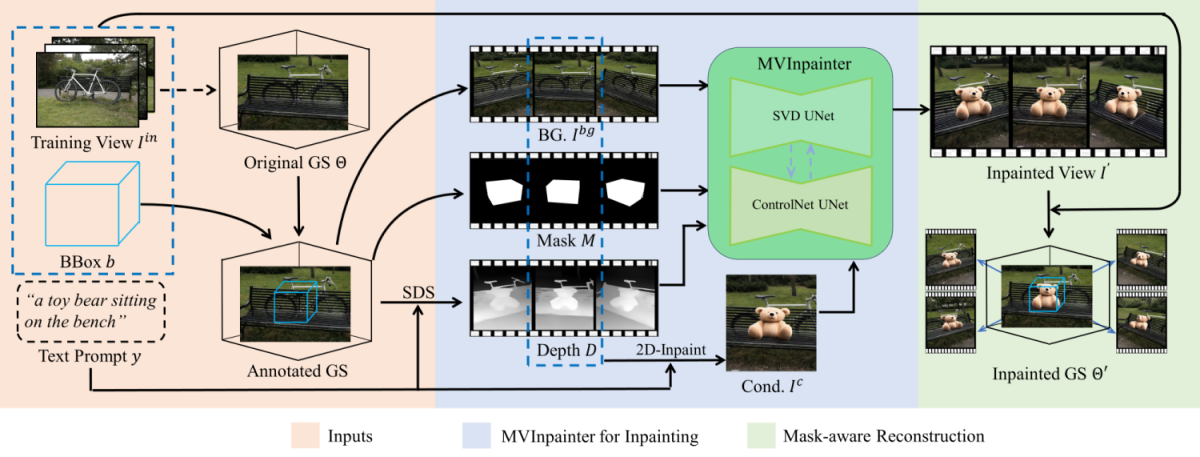}
    \caption{\textbf{Method Framework.} The framework is divided into three main parts: Inputs, MVInpainter for Inpainting, and Mask-aware Reconstruction. Initially, the inputs include the original Gaussian scene \(\Theta\), a bounding box (BBox) \(b\), and a text prompt \(y\). For the MVInpainter to perform inpainting, background images \(I^{bg}\), masks \(M\), and depth maps \(D\) are first derived. Using these inputs, along with the conditioning input \(I^c\), the MVInpainter generates consistent inpainted views \(I'\). Finally, in the Mask-aware Reconstruction phase, the inpainted Gaussian scene \(\Theta'\) is reconstructed using both inpainted and original training views for novel view synthesis, guided by a mask derived from the BBox \(b\).}
    \label{f_pipe}
\end{figure*}

\begin{figure}
    \centering
    \includegraphics[width=.95\columnwidth]{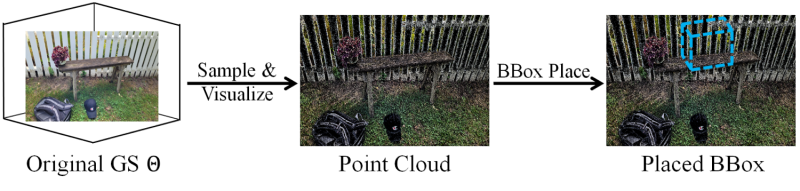}
    \caption{\textbf{Bounding Box Placement.} The placement of the bounding box enables the user to freely define the editing region in a convenient and intuitive manner.}
    \label{f_bbox}
\end{figure}

We aim to create new objects within a 3D scene based on a user-provided text prompt that specifies the target object. Additionally, the user can indicate the insertion region interactively. To achieve this, we have proposed a framework comprising three main components, as shown in Fig. \ref{f_pipe}:

\noindent\textbf{Inputs.} 
The inputs to our method include a pre-trained 3D Gaussian \(\Theta\) representing a 3D scene, a text prompt \(y\) describing the target object for insertion, and a 3D bounding box (BBox) \(b\) indicating the insertion region. We illustrate the BBox creation process in Fig. \ref{f_bbox}. First, we sample a point cloud from \(\Theta\). This point cloud is then visualized using 3D software. Next, we place a cube to represent the BBox. Users can easily translate, rotate, and scale the BBox to define the editing region.

Generating a new object from scratch without any shape reference poses a significant challenge in maintaining view consistency. To address this, we propose using SDS optimization \cite{poole2022dreamfusion,tang2023dreamgaussian} to provide a coarse geometry prior. Specifically, we generate a coarse geometry based on the text prompt \(y\) within the original 3D Gaussian \(\Theta\). We refer to this optimized 3D Gaussian with the coarse geometry prior as a coarse model \(\Theta_c\). It is important to note that, unlike previous works \cite{dai2024go, chen2024mvip, bartrum2024replaceanything3d}, we do not use the SDS optimization results as the final output. Instead, we use them solely as a coarse geometry prior, since SDS optimization can lead to poor visual results, particularly issues with over-smoothness and high saturation.

\noindent\textbf{MVInpainter for Inpainting.} 
We want the MVInpainter to capture the background from the original scene, the editing region from the BBox, and the geometry prior from the coarse 3D Gaussian model \(\Theta_c\), across a range of coherent viewpoints. To achieve this, we start by defining a circular trajectory revolving around the vertical axis of $b$. From this trajectory, we uniformly sample $n$ viewpoints, denoted as $\{C^{edit}_{i}\}_{i=1}^{n}$, where all cameras from these viewpoints are directed towards the center of $b$. For each editing viewpoint, we render the background image $I^{bg}$ based on the original Gaussian $\Theta$, generate the 2D editing mask $M$ corresponding to the BBox $b$, and produce a depth reference image $D$ derived from the coarse Gaussian ${\Theta}_c$. The MVInpainter takes the image sequences \(\{I^{bg}_{i}\}_{i=1}^{n}\), \(\{M_{i}\}_{i=1}^{n}\), \(\{D_{i}\}_{i=1}^{n}\), and the text prompt \(y\) as inputs, and produces view-consistent inpainting results \(\{I'_{i}\}_{i=1}^{n}\) that match the user requirements described by \(y\). 

In our approach, the MVInpainter consists of two branches: a multi-view diffusion module (MVD), inspired by the recent Stable Video Diffusion (SVD) \cite{blattmann2023stable}, and a ControlNet-based condition injection module, adapted from ControlNet \cite{zhang2023adding}.

As an image-to-video model, SVD generates a coherent set of image sequences based on an input image, known as the conditioning image. To adapt SVD into our MVD for consistent multi-view generation, we employ a depth-guided 2D inpainting method \cite{zhang2023adding, rombach2022high} to produce the conditioning image \(I^{c}\),
using the text prompt \(y\), background image $I^{bg}$, mask $M$, and depth map $D$ sampled from the central viewpoint of the inpainting trajectory. This \(I^{c}\) serves as the conditioning input for the fixed SVD. However, although the SVD excels at denoising video frames, our goal is to generate sparse yet consistent multi-view inpainting outcomes. We have noted that when relying solely on $I^{c}$ and the SVD, our MVD falls short in producing visually pleasing and camera-aware results.
Therefore, we propose including a ControlNet-based condition injection module. This module takes the depth \(D\), background \(I^{bg}\), and the mask \(M\) as inputs. It helps guide the inpainting for foreground content, manage the appearance of the scene background, and control the trajectory of the generated multi-view outputs. As a key component of our overall framework, we provide more details about our MVInpainter in Sec. \ref{MVD}.

\noindent\textbf{Mask-aware Reconstruction.} We then use the inpainting results \(\{I'_{i}\}_{i=1}^{n}\) to reconstruct a 3D Gaussian \(\Theta'\), incorporating the target object for novel view synthesis. Given that our MVInpainter, based on fixed SVD, is limited to generating only 14 views of the inpainting results, the sparse inpainted views usually can not provide comprehensive supervision for the complete scene background during the reconstruction of the edited scenes. Consequently, when rendering novel view images that fall outside the distribution of inpainted views, floating and noisy artifacts may arise. To address this, we introduce a mask-aware finetuning strategy to reconstruct the edited scenes by leveraging both the generated inpainted views and the input views of the original 3D Gaussian. This approach ensures sufficient supervision for the entire 3D scenes, leading to improved reconstruction quality of the edited scenes. More details are provieded in Sec. \ref{MAR}.

\section{MVInpainter and Mask-aware Reconstruction}\label{MVInpainter}

\begin{figure*}
	\centering
	\includegraphics[width=.9\textwidth]{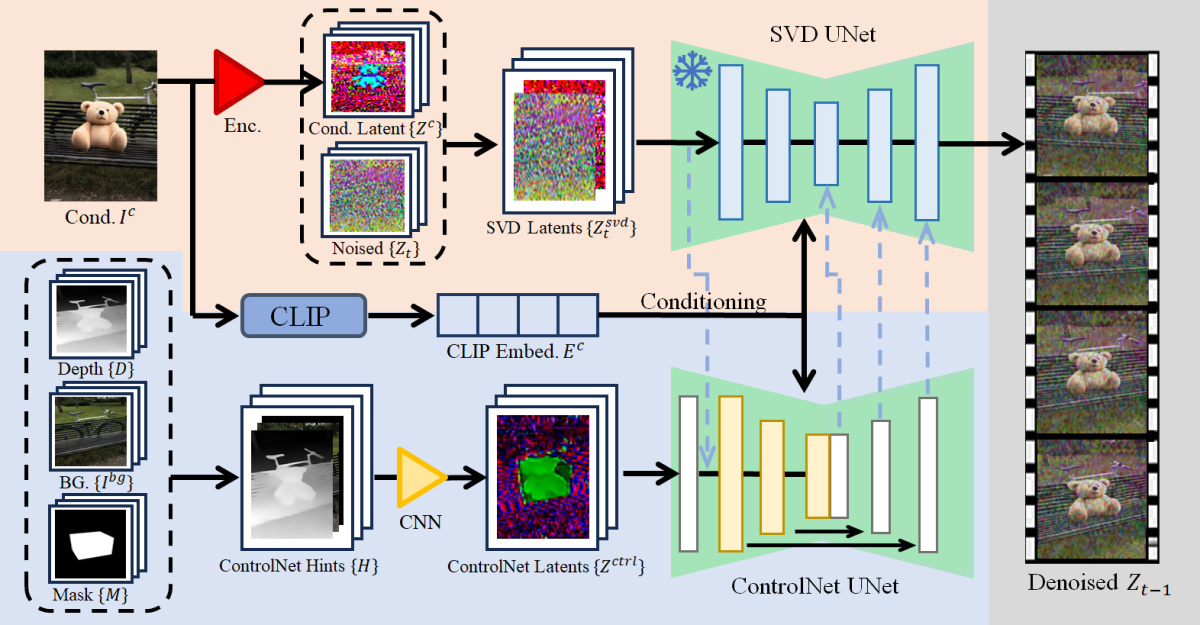}
	\caption{\textbf{MVInpainter Pipeline.} The MVInpainter integrates a multi-view diffusion module (MVD) and a ControlNet-based condition injection module to achieve view-consistent inpainting across multiple viewpoints. The MVD, adapted from the Stable Video Diffusion (SVD) model, generates consistent multi-view outputs conditioned on an image \(I^{c}\), while the ControlNet module refines the inpainting by guiding the foreground content, managing the background appearances, and controlling the trajectory of the generated outputs conditioned on the depth \(D\), the background $I^{bg}$, and the mask \(M\).}
	\label{f_svd}
\end{figure*}


In this section, we first delve deeper into the details of our MVInpainter, which comprises an MVD module and a ControlNet-based condition injection module (Sec.~\ref{MVD}), as illustrated in Fig.~\ref{f_svd}.
We then explain the technical details of our mask-aware reconstruction method (Sec.~\ref{MAR}).

\subsection{MVInpainter}\label{MVD}

As depicted in the upper part of Fig. \ref{f_svd}, our MVD module is built on the SVD framework, which is an image-to-video model fine-tuned from the Stable Diffusion Model. The SVD model takes a conditioning image, \(I^{c}\), as its input.
First, \(I^{c}\) is processed by the pre-trained VAE encoder of the SVD, generating an output that is duplicated \(n\) times to create a conditioning latent sequence \(\{Z^{c}_{i}\}_{i=1}^{n}\). These latent representations are then concatenated along the channel dimension with the noisy latent state inputs \(\{Z_{t,i}\}_{i=1}^{n}\) at time step \(t\), forming the input latent sequence \(\{Z^{svd}_{t,i}\}_{i=1}^{n}\) for the SVD UNet. Simultaneously, the CLIP embedding of \(I^{c}\), denoted as \(E^{c}\), is extracted and used as the conditioning input to provide semantic guidance. Finally, these inputs are fed into the SVD UNet, where they undergo denoising through the diffusion process.

While SVD is capable of producing coherent video sequences from conditioning images, the trajectory and content of the resulting videos are often unpredictable. The generated frames may not align with the intended viewpoints, and the backgrounds can diverge from the original scene. In contrast, our goal is to generate multi-view inpainting with specific camera trajectories, consistent foreground content, and preserved backgrounds.
To accomplish this, we introduce a ControlNet-based condition injection module. This module integrates additional reference information into the SVD, enabling controlled and more predictable multi-view generation.

As illustrated in the lower part of Fig. \ref{f_svd}, our ControlNet module receives inputs comprising depth maps \(\{D_{i}\}_{i=1}^{n}\), 2D editing masks \(\{M_{i}\}_{i=1}^{n}\), and background images \(\{I^{bg}_{i}\}_{i=1}^{n}\). Following \cite{rombach2022high}, we initially multiply $I^{bg}$ by $M$, then concatenate the result with $M$ and $D$ along the channel dimension to create ControlNet Hints \(\{H_{i}\}_{i=1}^{n}\). Subsequently, these hints are input into a trainable CNN downsampler for size adjustment, producing the latent sequence \(\{Z^{ctrl}_{i}\}_{i=1}^{n}\) for the ControlNet UNet. This latent sequence encapsulates reference information, including the rough geometries of the target object, background appearances for each viewpoint, and the editing areas on each viewpoint. In accordance with \cite{zhang2023adding}, these inputs undergo processing through a zero-initialized convolution layer before being combined with the latent input from SVD, \(\{Z^{svd}_{t,i}\}_{i=1}^{n}\). The combined inputs are then processed by the encoder and middle blocks within ControlNet. These blocks mirror the structure of their counterparts in the pre-trained SVD, with weights initialized by direct replication. The outputs from ControlNet's encoder and middle blocks pass through several zero-initialized convolution layers before being combined with the skip-connection inputs of the corresponding SVD blocks, injecting control signals into the SVD model.

By integrating our MVD and ControlNet injection modules, essential references such as the backgrounds of each viewpoint, the editing regions, and the shape prior of the target object are incorporated into the SVD generation process. This combination results in controlled, multi-view outcomes \(\{Z_{t-1,i}\}_{i=1}^{n}\). After denoising, the final outputs \(\{Z_{0,i}\}_{i=1}^{n}\) are decoded via the SVD decoder, producing the view-consistent inpainted results that align with the user requirements described by \(y\). To ensure the faithful preservation of the scene background in inpainted images, we further employ a video segmentation model \cite{cheng2023segment,kirillov2023segment,liu2023grounding} to segment the target objects and integrate them with the original background, resulting in the final inpainting results, \(\{I'_{i}\}_{i=1}^{n}\). Subsequently, these images are leveraged to optimize the 3D Gaussian, leading to the edited 3D scene.

\subsection{Mask-aware Reconstruction}
\label{MAR}

Here, we delve into the technical details of our mask-aware reconstruction method.
First, for the \(u\)-th pixel at the \(i\)-th edited viewpoint, we compute the rendered color using the following formula:
\begin{equation}
    I^{r}_{i,u} = \sum\limits_{j \in N_p} c_j \alpha_j \prod\limits^{j-1}_{k=1} (1 - \alpha_k).
\end{equation}
where \(N_p\) is a set of sampling points along the camera ray \(r\), and \(c_j\) and \(\alpha_j\) represent the color and opacity corresponding to sampling point \(j\), respectively. Subsequently, we employ the reconstruction loss \cite{kerbl20233d} to finetune the initial 3D Gaussian $\Theta$:
\begin{equation}
    L_{gs}(I^{r}, I^{gt}) = (1 - \lambda) \cdot |I^{r} - I^{gt}| + \lambda \cdot (1 - SSIM(I^{r}, I^{gt})),
\end{equation}
where \(I^{gt}\) is the ground truth image corresponding to \(I^{r}\), and \(\lambda\) is a constant weight set to 0.2 following \cite{kerbl20233d}. 
To ensure that the background remains completely unchanged before and after finetuning, we enhance constraints on the background area by performing reconstruction optimization on the unmasked regions in the training images, in addition to optimizing the edited viewpoints. This helps maintain the integrity of the background:
\begin{equation}
    L_{rec} = 
    \begin{cases}
        L_{gs}(I^{r}, I'), C \in \{C^{edit}\},\\
        L_{gs}(I^{r} \cdot (1 - M), I^{gt} \cdot (1 - M)), C \in \{C^{gt}\}.\\
    \end{cases}
\end{equation}
{where \(C^{gt}\) stands for the training set viewpoints of the original Gaussian. We use this reconstruction loss as our final optimization constraint.

Our mask-aware finetuning strategy significantly improves the reconstruction quality of 3D Gaussians, minimizing floating and noisy artifacts in novel view synthesis. This approach ensures high-quality results even with sparse inpainted views.

\section{Experiments}

\subsection{Experimental Setup}\label{imp}

\noindent\textbf{Training Dataset.} 
We utilize the Wild-RGBD dataset \cite{xia2024rgbd} to train our MVInpainter. The vanilla dataset comprises about 7,000 collections of single-object-centric scenes spanning 46 common object categories. Each scene in the dataset is captured from around 300 viewpoints along a 360-degree circular trajectory, providing detailed scene appearances and precise object masks at each viewpoint. To improve data quality and ensure its suitability for our generation task, we implement a filtering process that excludes 13 categories with nearly-flat geometry from the original data. Additionally, we carefully select 50 scenes from each of the remaining categories to construct our dataset. In total, our dataset comprises approximately 1,600 scenes, covering a wide range of diverse categories. To create training samples from these selected scenes, we first sample a circular trajectory that matches the range of our inpainting views in each scene. Subsequently, we evenly sample $n$ viewpoints along this trajectory to extract the corresponding observed images and precise masks of the target objects. To eliminate the need for fine-grained masks during inference, we use dilation operations to generate square masks. Besides, we estimate a depth map \cite{ranftl2020towards} for each observed image, serving as a reference for inpainting.
Ultimately, the observed image sequence, square masks, and depth references of a scene collectively constitute a training data sample that can be used to train our MVInpainter.

\begin{figure*}
	\centering
	\includegraphics[width=.95\textwidth]{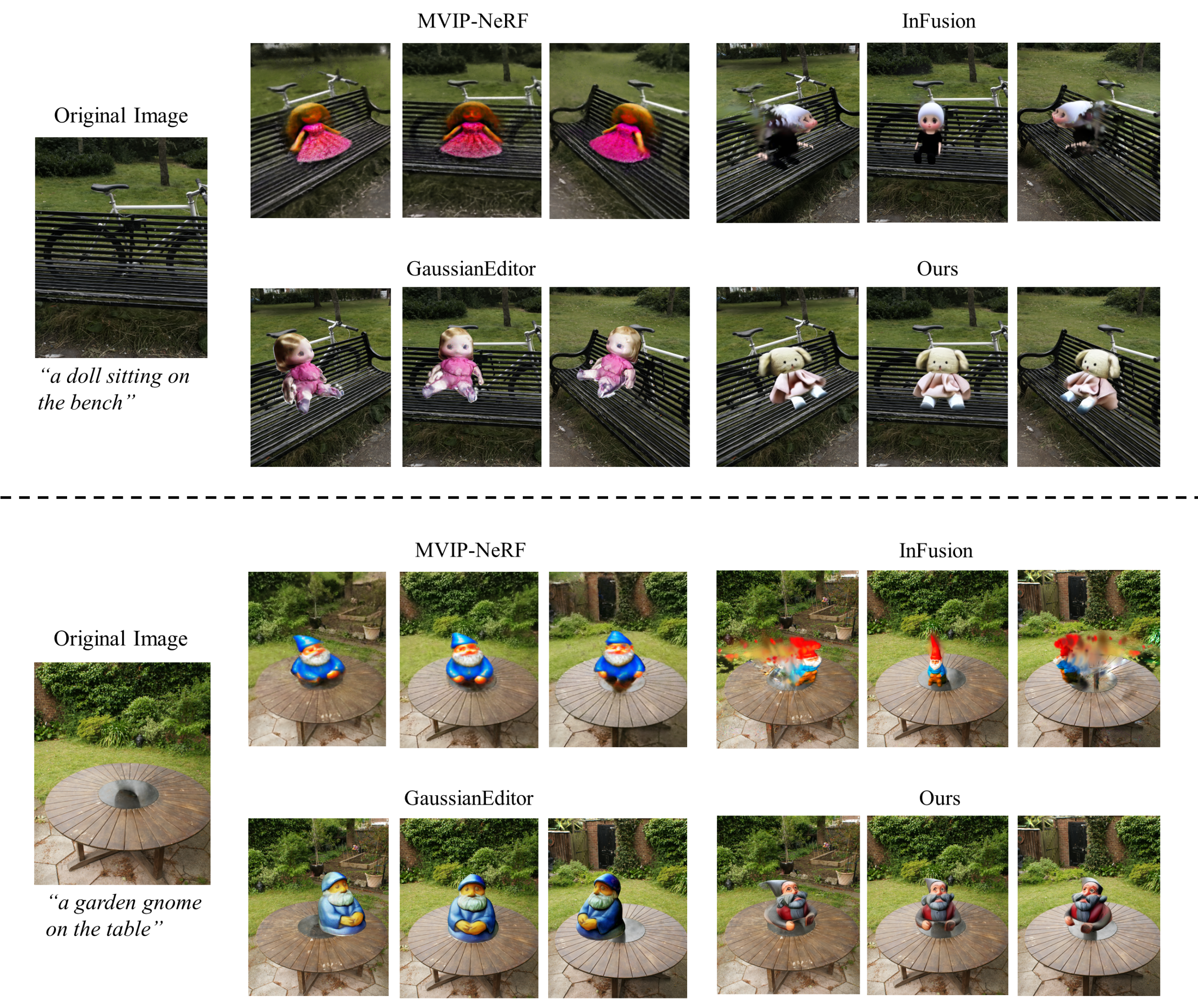}
	\caption{\textbf{Qualitative Comparison with Other State-of-the-art Methods.} Our approach directly generates view-consistent appearances across multiple viewpoints, bypassing SDS optimization and ensuring harmonious integration of generated objects with the scene from various angles. As a result, it achieves the most authentic generative object insertion.}
	\label{f_sota}
\end{figure*}

\begin{table}[width=.9\columnwidth]
\caption{\textbf{Quantitative Comparison with State-of-the-art Methods.} Compared to other models, our approach excels in three key aspects: the consistency between editing results and textual descriptions, the effectiveness of edits, and the authenticity of the editing effects.}\label{tab_sota}
\begin{tabular}{cccc}
\toprule
 & CTIS↑ & DTIS↑ & MUSIQ↑ \\
\midrule
{\bf Ours} & {\bf 0.2977} & {\bf 0.2033} & {\bf 71.391} \\
MVIP-NeRF & 0.2682 & 0.1054 & 38.177 \\
InFusion & 0.2724 & 0.1422 & 65.500 \\
GSEditor & 0.2893 & 0.1949 & 68.095 \\
\bottomrule
\end{tabular}
\end{table}

\begin{figure*}
	\centering
	\includegraphics[width=.99\textwidth]{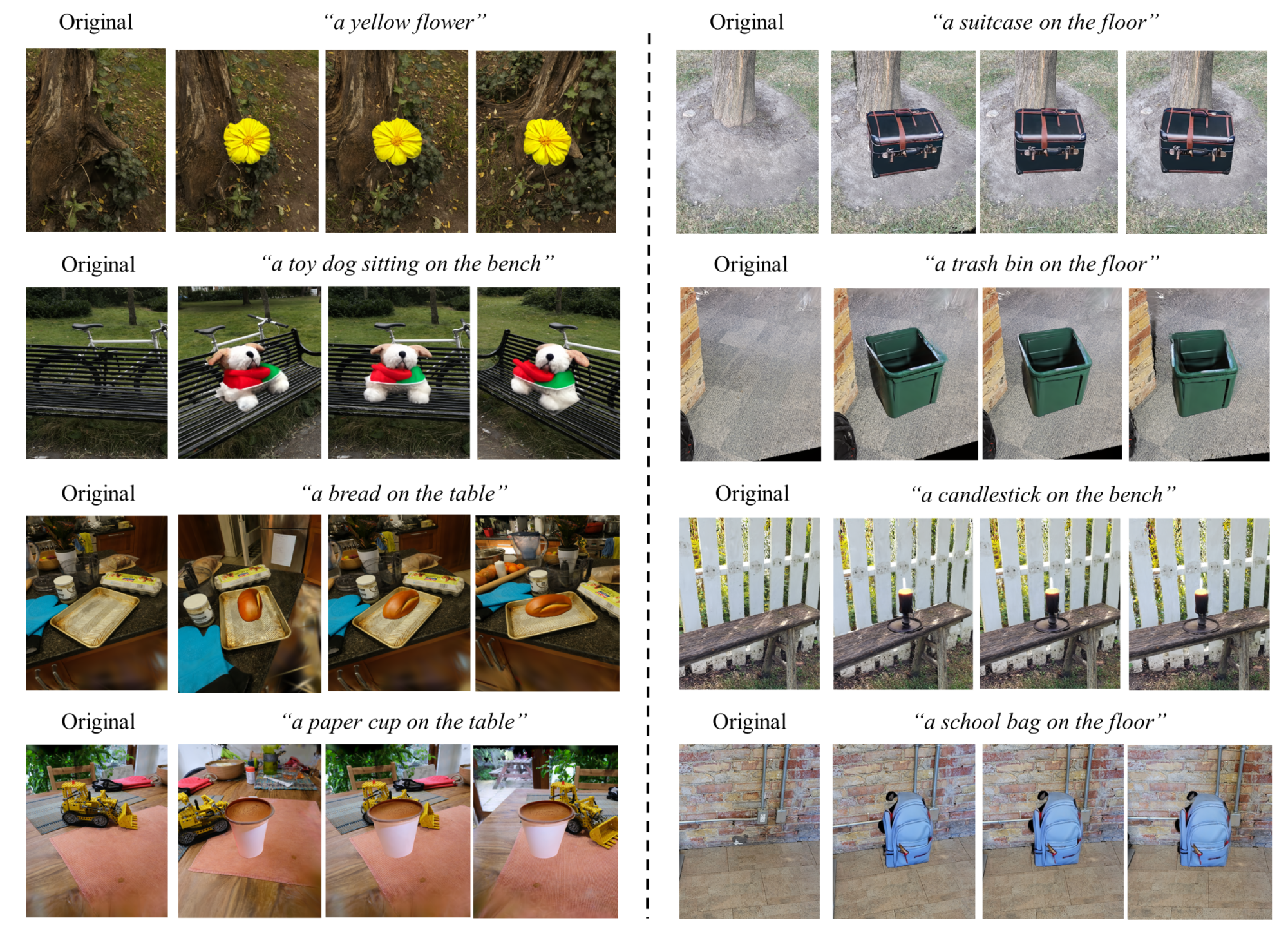}
	\caption{\textbf{Editing Results in Various Scenes.} Clearly, our model can achieve realistic and effective editing in diverse scenes.}
	\label{f_showcases}
\end{figure*} 

\begin{figure*}
	\centering
	\includegraphics[width=.95\textwidth]{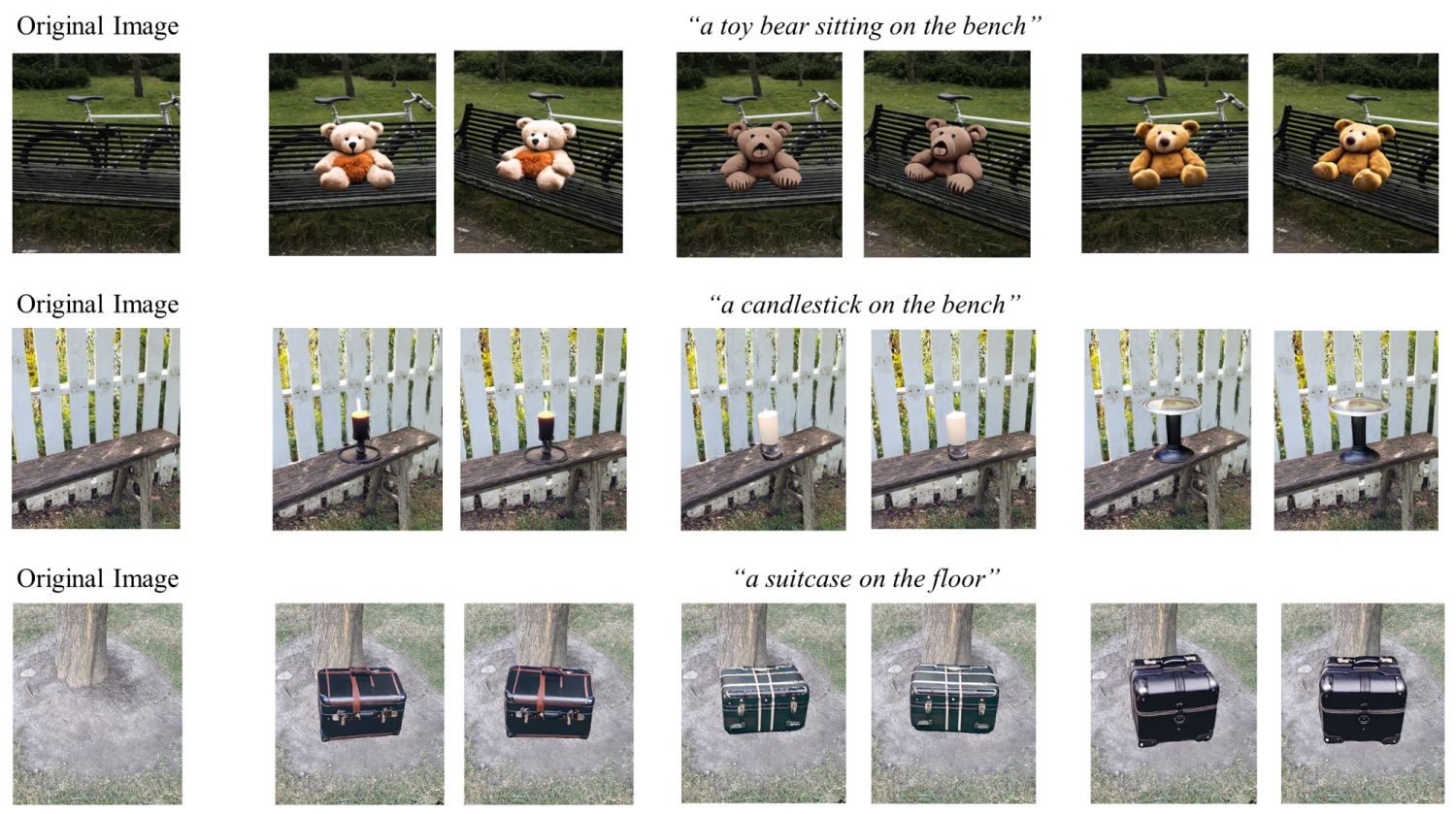}
	\caption{\textbf{Diverse Editing with the Same Setup.} Given the same text and editing region, our model can produce a variety of editing results, all of which harmonize with the scene and maintain a high level of realism. This demonstrates the strong generative capability of our approach.}
	\label{f_diverse}
\end{figure*}

\noindent\textbf{Evaluation Dataset.} To comprehensively evaluate the generation capabilities of relevant methods across diverse scenes, we curated facing-forward scenes from the SPIn-NeRF dataset \cite{mirzaei2023spin} and unbounded scenes from the MipNeRF-360 dataset \cite{barron2022mip} for our experimental analysis. In total, we use nine scenes (\textit{tree}, \textit{corner}, \textit{bench}, \textit{office}, \textit{bicycle}, \textit{garden}, \textit{kitchen}, \textit{stump}, \textit{counter}) to generate our editing samples.

\noindent\textbf{Baseline Methods.} We compare our model with three state-of-the-art methods, including one SDS-based approach, MVIP-NeRF \cite{chen2024mvip}, and two 3D reconstruction-based inpainting methods for object insertion: InFusion \cite{liu2024infusion} and GaussianEditor \cite{chen2024gaussianeditor}. We use the officially released code to conduct the comparison experiments.

\noindent\textbf{Metrics.} Following InseRF \cite{shahbazi2024inserf} and InFusion \cite{liu2024infusion}, we integrate CLIP Text-Image Similarity (CTIS) to evaluate the consistency between editing results and input texts, and utilize Directional Text-Image Similarity (DTIS) to assess the effectiveness of edits. Here, CTIS computes the cosine similarity between the CLIP \cite{radford2021learning} embeddings of the text prompt and the rendered images of the edited scene. DTIS measures the similarity of the changing directions between the image and text CLIP embeddings. Additionally, we further employ MUSIQ \cite{ke2021musiq}, a non-reference image quality assessment model, to evaluate the authenticity of our edited results \cite{mirzaei2023reference}.

\begin{figure}
	\centering
	\includegraphics[width=.95\columnwidth]{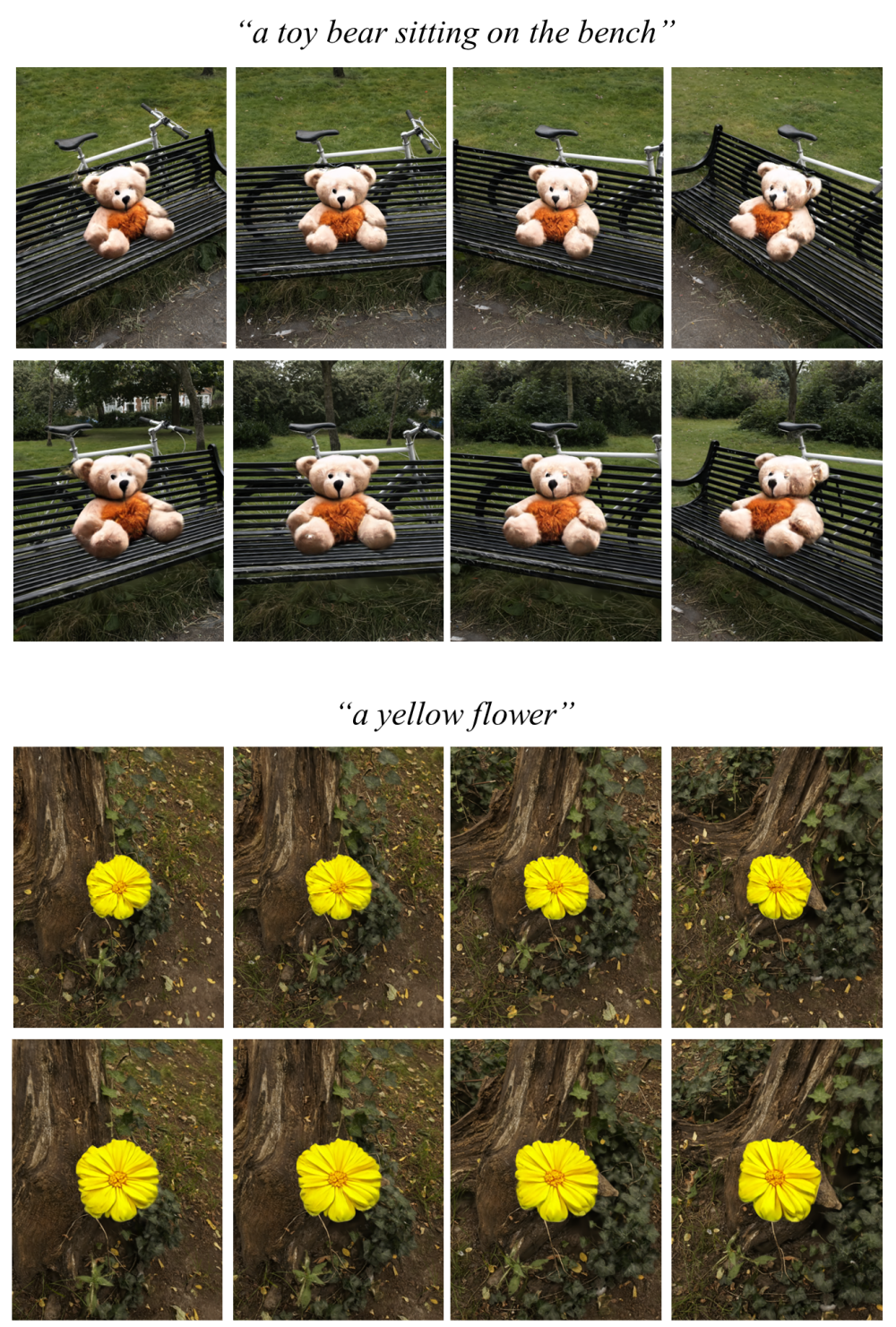}
	\caption{\textbf{3D-consistent Editing across Diverse Viewpoints.} The 3D content generated by our model maintains high quality and strict view-consistency across a wide range of observation angles.}
	\label{f_views}
\end{figure}

\begin{figure}
	\centering
	\includegraphics[width=.95\columnwidth]{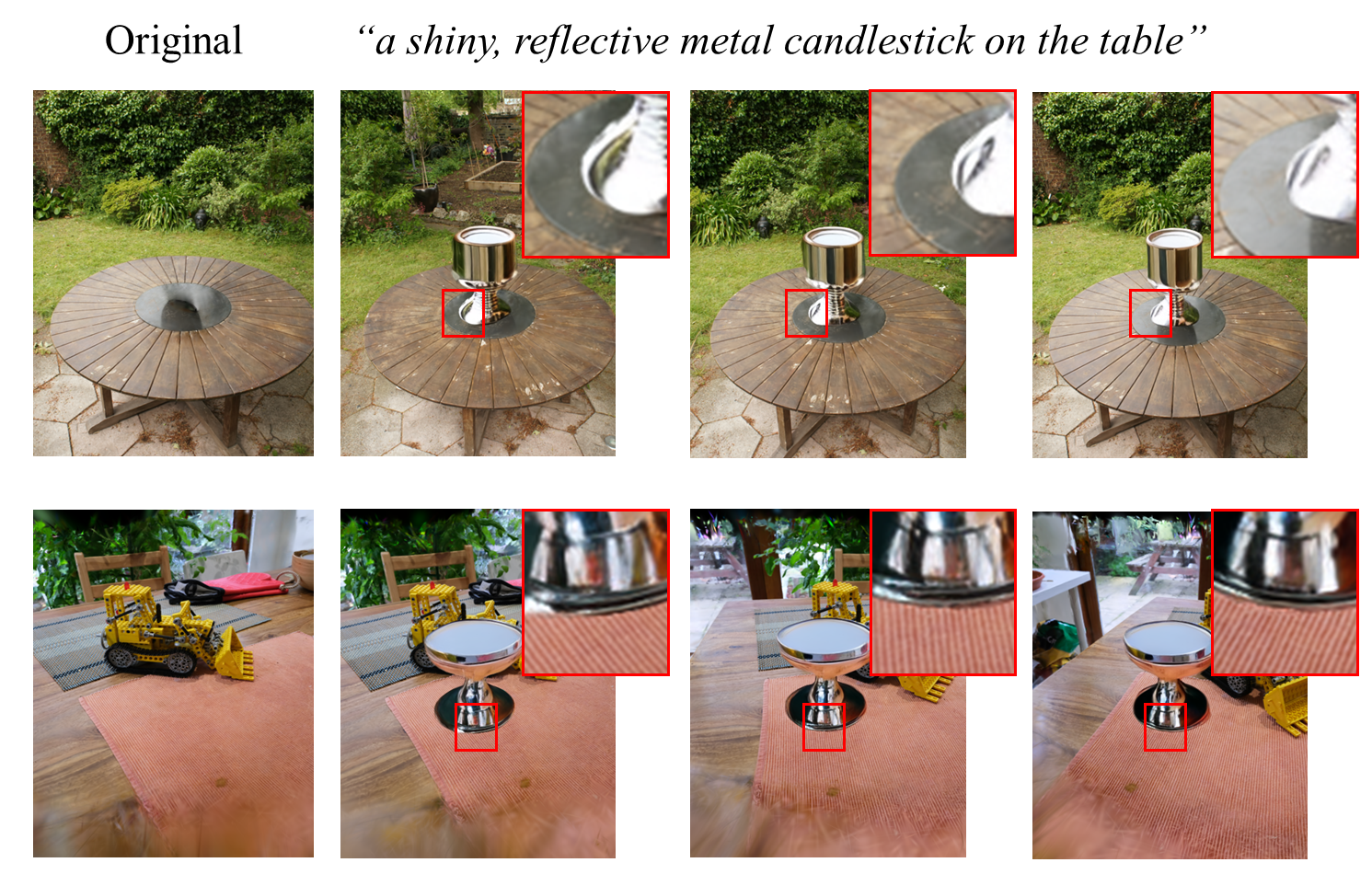}
	\caption{\textbf{Generation of Reflective Objects.} Our model also supports the generation of reflective objects. The generated surfaces exhibit prominent view-dependent effects, such as specular reflections.}
	\label{f_reflect}
\end{figure}

\noindent\textbf{Training Details.} We train our MVInpainter following a similar optimization process as SVD \cite{blattmann2023stable}. Specifically, we synchronize our network parameters with those of the SVD model \cite{blattmann2023stable} and conduct 50 epochs of optimization. The training process is performed on four A6000 GPUs, taking approximately 30 hours to complete. For each scene editing task, we initially optimize the 3D Gaussian of the original scene by executing 5000 iterations of SDS optimization to obtain a coarse geometry prior, following the parameter configurations specified in DreamGaussian \cite{tang2023dreamgaussian}. Once the coarse geometry and other necessary inputs are prepared, we respectively sample $n=14$ inpainting views on the left and right half of a 120-degree circular trajectory. Subsequently, we leverage the trained MVInpainter to generate the inpainting results for these viewpoints. With the inpainted images, we apply our mask-aware reconstruction strategy, engaging in 30,000 iterations of optimization to achieve the final edited scene. The parameters for this optimization step, including the learning rate, interval for adaptive density control, and densification threshold, are configured in line with the official Gaussian Splatting code \cite{kerbl20233d}. The editing process for a single scene can be completed in approximately one hour on a single GPU.

\subsection{Comparisons}

We conduct quantitative and qualitative comparative experiments between our proposed method and the baseline approaches using curated datasets. The results are displayed in Table \ref{tab_sota} and Fig.~\ref{f_sota}. Our approach outperforms other models in three key areas: maintaining consistency between editing results and textual descriptions, delivering effective edits, and achieving realistic editing effects.

Specifically, MVIP-NeRF relies on SDS optimization for object generation. However, in SDS optimization, the update direction at each iteration demonstrates significant randomness \cite{liang2024luciddreamer}. With these diverse update directions applied to the same 3D model, the final results produced by MVIP-NeRF tend to be overly smooth due to the averaging effect, as depicted in Fig. \ref{f_sota}. Moreover, SDS optimization requires coupling with a large conditional guidance scale, leading to oversaturated results in content generation. In contrast, our approach involves directly creating 3D-consistent inpainting images from various viewpoints and reconstructing the target object using these images. Such strategy allows us to generate high-quality 3D content without the drawbacks associated with SDS optimization.

On the other hand, InFusion performs 2D inpainting on a single view and then projects the inpainted appearance back into 3D based on estimated depth. Nonetheless, since the projected 3D appearance is solely derived from a single-view observation, it can lead to incomplete shapes of the target object. This, in turn, results in fragmented side views of the created content, as depicted in the \textit{doll} case. Conversely, GaussianEditor first reconstructs the target object from the inpainted image and then places the reconstructed model in 3D to ensure shape completeness. However, due to the side view generation of the reconstructed object does not involve inpainting, it fails to ensure that the created side views of the target object blend seamlessly with the scene, as demonstrated by the black patches on the left side view in the \textit{doll} case. Additionally, due to the absence of depth information in single-view inpainting, both InFusion and GaussianEditor rely on inaccurate depth estimation for object placement. Consequently, as depicted in the \textit{gnome} case, InFusion may project parts of the gnome too deeply, leading to floating artifacts. Meanwhile, even after meticulous manual adjustments, GaussianEditor fails to position the gnome accurately, making it floating above the table. In contrast to these methods, our multi-view inpainting approach excels in accurately placing created objects and seamlessly blending their appearance with the scene background across various views, leading to superior editing results.

In addition to outperforming in qualitative comparison, our method also stands out in quantitative evaluation. As shown in Table \ref{tab_sota}, our method generates effective editing that best align with the text prompts across multiple viewpoints, achieving top scores in the CLIP-relevant metrics CTIS and DTIS. Additionally, the objects generated by our method exhibit a highly realistic appearance and blend seamlessly with the scene background, leading to superior performance in the MUSIQ metric that assesses the authenticity of rendered images.

\subsection{Editing Results}

To further illustrate the editing capabilities of our method, we present a wider range of editing results across an increased number of test samples in Fig.~\ref{f_showcases}. It is clear that our method excels in producing results that seamlessly integrate with indoor and outdoor scenes, even under varying lighting conditions. Notably, in cases such as the \textit{flower} and \textit{dog} examples, our method exhibits a remarkable ability to preserve view-consistency when creating objects with intricate shapes, underscoring the effectiveness of our approach in achieving high-quality edits across a diverse array of scenes.

Additionally, as shown in Fig.~\ref{f_diverse}, our method demonstrates the capability to generate a diverse set of editing outcomes under uniform editing conditions. For instance, in the \textit{candlestick} case, our method creates candle holders of different sizes, colors, and materials, some lit while others remain unlit. Similarly, in the \textit{bear} case, the first set of examples features a brighter appearance, while the second set exhibits a darker sheen. Nevertheless, both variations seamlessly blend with the overall scene, appearing highly realistic and showcasing the versatility of our approach.

Furthermore, as depicted in Fig.~\ref{f_views}, the editing results produced by our model maintain a high level of realism and viewpoint consistency across a broad range of views covered by our inpainting. This highlights that within this coverage, the reconstructed 3D objects consistently display well-reconstructed appearances when viewed from arbitrary angles.
Even in scenarios involving complex shapes, such as the \textit{flower} case, our method consistently generates realistic and harmonious 3D results across various viewing angles, whether close or distant, high or low, highlighting the robustness of our approach.

Besides, as depicted in Fig. \ref{f_reflect}, our model also facilitates the generation of reflective objects. For instance, in the \textit{candlestick} case in the top row, the central portion of the zoomed-in area does not have a highlight in the initial viewpoint. Upon transitioning to the second viewpoint, a leaf-shaped highlight becomes visible. Subsequently, in the third viewpoint, this highlight expands and blends with the surrounding highlights. This outcome showcases the efficacy of our model in producing objects with view-dependent effects.

\begin{figure*}
	\centering
	\includegraphics[width=.95\textwidth]{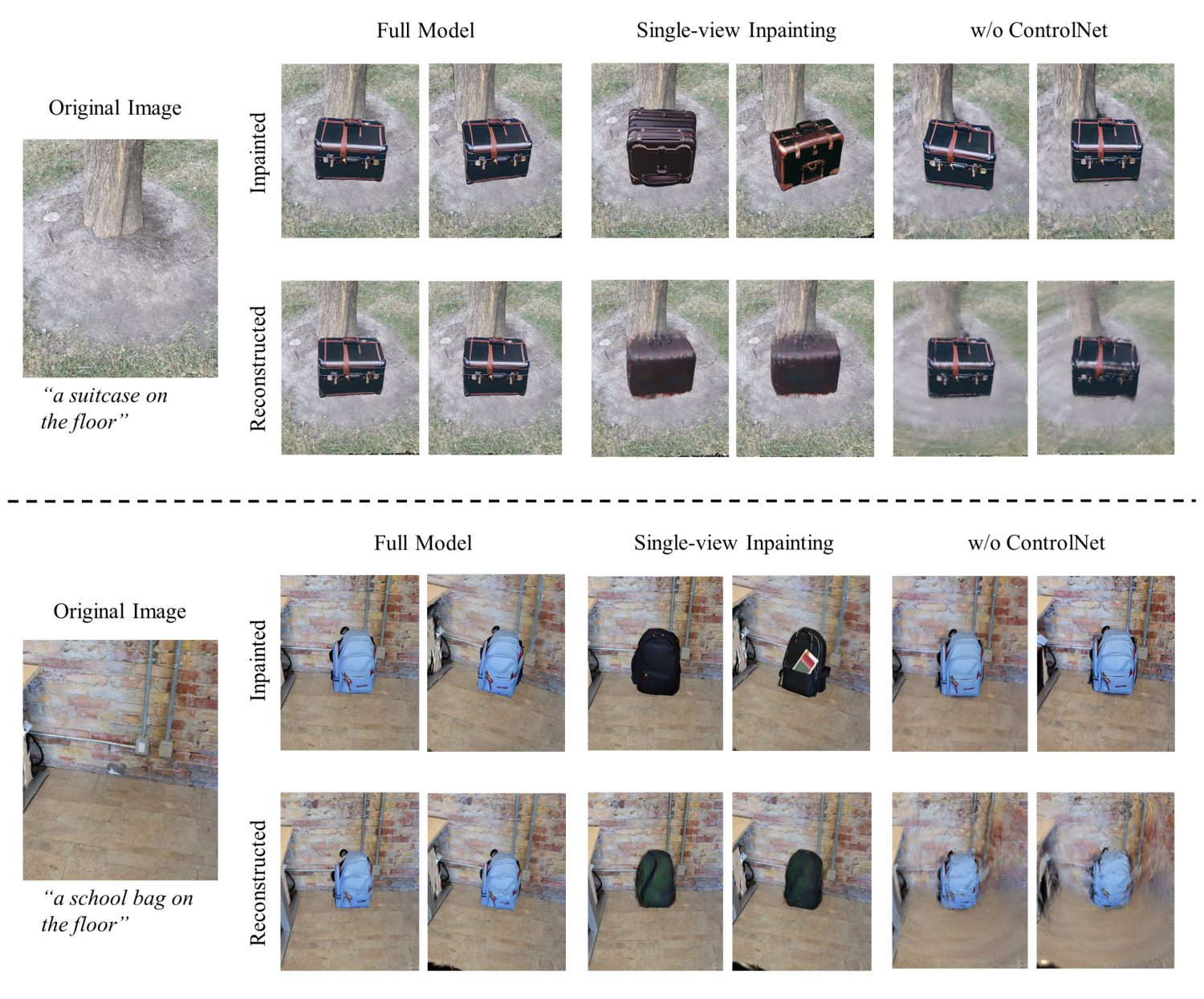}
	\caption{
\textbf{Ablation on Multi-view Inpainting and ControlNet Module.}
Directly applying 2D inpainting methods for editing across various viewpoints can not guarantee view-consistency, leading to noticeable artifacts in the reconstructed 3D content. Without incorporating our ControlNet module, editing directly using SVD leads to generated content that does not align with the corresponding viewpoints, making it difficult to reconstruct the editing outcomes effectively. In contrast, our full model can produce editing results that are view-consistent and match the corresponding camera positions, thereby generating high-quality 3D content.}
	\label{f_ablation}
\end{figure*}

\begin{figure*}
	\centering
	\includegraphics[width=.95\textwidth]{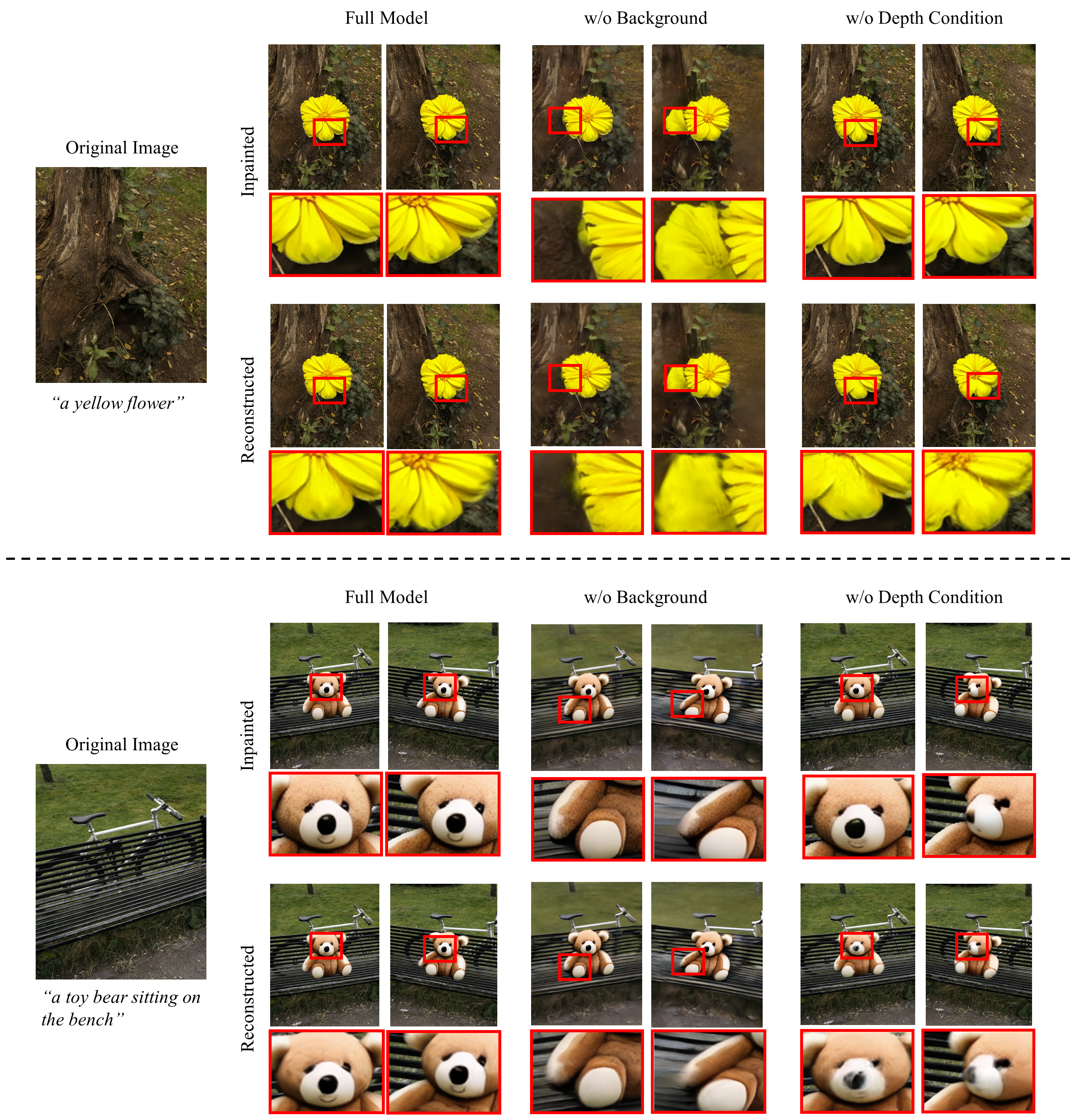}
	\caption{
\textbf{Ablation on Input Conditions.}
When background inputs are absent, the model struggles to analyze the interaction between the foreground and background, leading to artifacts at the boundary of the editing area. On the other hand, when lacking depth reference inputs, the appearance generated by the model sometimes fails to guarantee strict 3D-consistency, resulting in blurriness in the reconstruction. In contrast, our full model can produce editing results that interact well with the background and maintain view-consistency.}
	\label{f_ablation_ctrl}
\end{figure*}

\begin{figure*}
	\centering
	\includegraphics[width=.95\textwidth]{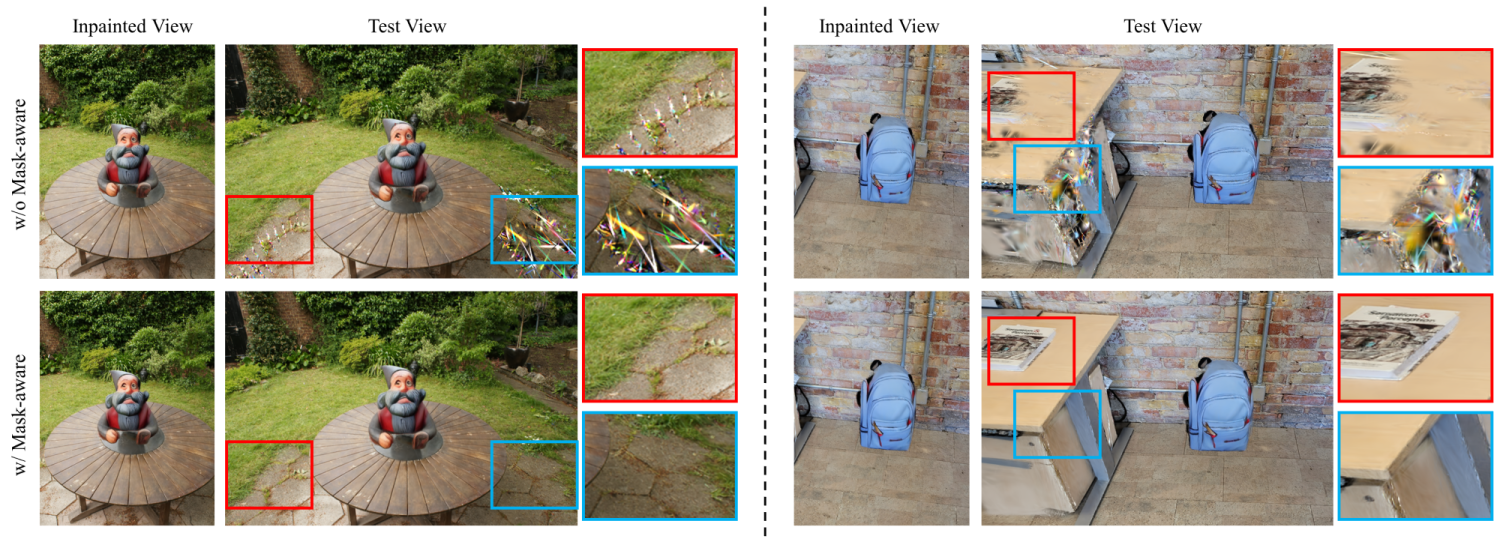}
	\caption{
\textbf{Ablation on Mask-aware Reconstruction.}
In the absence of mask-aware reconstruction, relying solely on sparse inpainted views is insufficient to provide adequate supervision for the complete background of the original scene. This deficiency leads to artifacts in the reconstructed background when viewed from angles beyond those of the inpainted views. In contrast, our mask-aware reconstruction utilizes both the inpainted view and training views to constrain the reconstruction process, ensuring the correct reconstruction of the scene background during the reconstruction of the target object.}
	\label{f_ablation_mask-aware}
\end{figure*}

\subsection{Ablation Study} 

\noindent\textbf{Multi-View Inpainting.} To investigate the superiority of using MVInpainter for multi-view inpainting, we carry out a comparative experiment by replacing MVInpainter with a single-view inpainting model \cite{zhang2023adding, rombach2022high} and independently creating inpainted content for each viewpoint. The qualitative outcomes are depicted in Fig. \ref{f_ablation}. Evidently, the single-view inpainting model can not provide constraints on 3D consistency, resulting in divergent objects being generated in different views. For example, in the \textit{suitcase} case, the single-view inpainting method produces dissimilar suitcases in two separate views, leading to blurry and inconsistent reconstructions. In contrast, our MVInpainter excels at producing view-consistent results across multiple viewpoints. For instance, it generates a consistently colored and shaped suitcase across different views in the \textit{suitcase} case, enabling view-consistent, high-quality generative object insertion. In addition, we further conduct a comparison to assess the view-consistency of the inpainted images by measuring the reconstruction error between the inpainted images and the reconstructed images. The detailed results can be found in Table \ref{abl}. Obviously, both quantitative and qualitative results demonstrate that the content generated by MVInpainter exhibits improved 3D consistency.

\noindent\textbf{ControlNet Module.} To validate the necessity of integrating the ControlNet Module into our pipeline, we conducted a comparison by removing the ControlNet Module from our framework and directly utilizing the pretrained SVD model for generation. The quantitative and qualitative results are presented in Table \ref{abl} and Fig. \ref{f_ablation}. As shown in the \textit{bag} case of Fig. \ref{f_ablation}, only by incorporating the ControlNet module can we generate proper inpainted images that precisely match the camera positions. Specifically, the content produced by single SVD shows minimal changes in appearance between the two viewpoints. However, the actual camera position undergoes a noticeable rotation between the views, leading to significant blurriness in the reconstructed results using this incomplete method. In contrast, our full model produces inpainting content on both viewpoints that align with the camera trajectory, ensuring that the generated content can be effectively reconstructed, thereby facilitating 3D content generation.

\begin{table}[width=.95\columnwidth]
\caption{\textbf{Ablation on the Superiority of Multi-view Inpainting, the Necessity of ControlNet Module, and the Effectiveness of Input Conditions.} We evaluate the view-consistency of the editing results by estimating the reconstruction error between inpainted images and the reconstructed images.}\label{abl}
\tabcolsep=0.08cm
\begin{tabular}{cccccc}
\toprule
 & {\bf Full} & 2D Inpaint & SVD & w/o BG. & w/o Depth \\
\midrule
PSNR↑ & {\bf 34.962} & 24.057 & 21.430 & 26.811 & 34.634 \\
SSIM↑ & {\bf 0.9778} & 0.9301 & 0.6634 & 0.8236 & 0.9776 \\
\bottomrule
\end{tabular}
\end{table}

\noindent\textbf{Input Conditions.} To validate the necessity of each input for our ControlNet module, we conducted a comparison between the results obtained using all inputs and those achieved using only a subset of inputs. Notably, the inclusion of the editing mask input is crucial for the training of MVInpainter. Therefore, it is not feasible to compare the results obtained without utilizing this input. The quantitative and qualitative comparisons are detailed in Table \ref{abl} and illustrated in Fig. \ref{f_ablation_ctrl}, respectively. In the absence of \textit{background} inputs, it becomes evident that the model struggles to capture accurate background environmental information across varying viewpoints. This limitation leads to the generation of noticeable ghosting artifacts along the boundary of the editing region. For example, in the \textit{flower} case's second viewpoint, an unnatural extra portion of the flower is synthesized when background inputs are absent. Conversely, our full model adeptly captures the environmental background appearance across diverse viewpoints, enabling accurate differentiation between the foreground and background of the image, thereby facilitating the generation of distinct new content. On the other hand, the lack of the \textit{depth condition} inputs results in inconsistencies in the generated shape details across viewpoints. For instance, in the \textit{flower} case, the MVInpainter generates three petals pointing towards the lower left, directly down, and lower right within the red box. However, in the second view of the results without depth input, both the middle and right petals have shifted significantly upwards, pointing towards the lower right and right, respectively. This shift in position between viewpoints introduces significant blurriness in the reconstructed results. In contrast, incorporating the depth reference ensures consistent shapes and positions of elements, enhancing clarity in the reconstructions.

\noindent\textbf{Mask-Aware Reconstrucion.} To illustrate the effectiveness of our mask-aware reconstruction technique, we conducted a comparison between the editing results with and without this setting. The relevant results are depicted in Fig. \ref{f_ablation_mask-aware}. It is evident that when relying solely on inpainted views for reconstruction, the limited range of viewpoints fails to offer adequate constraints for the scene background. This deficiency leads to visible artifacts at the boundary of the inpainted views' observation range. In contrast, the implementation of our mask-aware reconstruction leverages both the background of the training views and inpainted views to reconstruct edited results. This strategy ensures that the background of the target 3D Gaussian benefits from ample constraints derived from multiple viewpoints, consequently minimizing the occurrence of artifacts. In Table \ref{abl_mask}, we quantify the reconstruction error between the unmasked regions of the original and edited scene when utilizing or not utilizing mask-aware reconstruction. The results demonstrate that, the quality of the background reconstructed through mask-aware reconstruction surpasses the results achieved without employing this technique.

\begin{table}[width=.95\columnwidth]
\caption{\textbf{Ablation on Mask-aware Reconstruction.} Our full model can offer comprehensive constraints for the background of the original scene during the reconstruction of the target object, resulting in high-quality reconstruction of the scene background.}\label{abl_mask}
\begin{tabular}{ccc}
\toprule
 & PSNR↑ & SSIM↑ \\
\midrule
{\bf w/ Mask. Rec.} & {\bf 35.592} & {\bf 0.9610} \\
w/o Mask. Rec. & 29.838 & 0.9190 \\
\bottomrule
\end{tabular}
\end{table}

\section{Conclusion}

We present a novel approach for 3D generative object insertion. Unlike existing methods that rely on SDS optimization or single-view inpainting with reconstruction, our approach introduces a ControlNet in conjunction with a pretrained video diffusion model, resulting in the development of a multi-view inpainting diffusion model, MVInpainter. This model ensures consistent inpainting results across various viewpoints, enabling the high-quality generation of 3D-consistent objects that seamlessly integrate into the scene from multiple angles. Additionally, we propose a mask-aware 3D reconstruction technique to improve Gaussian Splatting reconstruction from sparse inpainted views. Extensive experiments demonstrate the effectiveness of our approach in ensuring consistent, realistic, and harmonious insertions, while generating diverse, high-quality 3D content.

\noindent\textbf{Limitations.} While our model excels at generative object insertion, it does have certain limitations.
One key challenge is the lack of publicly available sufficient training data, which limits the model's ability to support 360-degree content generation, as illustrated in Fig. \ref{f_360}. This limitation could be overcome as more training data becomes available in the future.
Another limitation is that the method cannot be directly extended to effective object removal, as shown in Fig. \ref{f_removal}.
The main challenge is that the newly patched background often fails to align seamlessly with the original scene, resulting in reconstruction artifacts. Implementing distinct inpainting strategies for void areas and background regions within the editing zone could help address this issue, and we plan to explore this approach in future work.
In addition, our current model has difficulty handling shadows from inserted objects. To ensure the newly-generated background remain unchanged, we extract the target object after inpainting and reintegrate it into the original background for subsequent reconstruction. However, as depicted in Fig. \ref{f_shade}, the segmentation model sometimes overlooks shadows, resulting in reconstructions without them. Manual adjustments to the mask or using advanced segmentation methods could resolve this issue, as demonstrated by the manually segmented cases on the right side of Fig. \ref{f_shade}.

\begin{figure}
	\centering
	\includegraphics[width=.95\columnwidth]{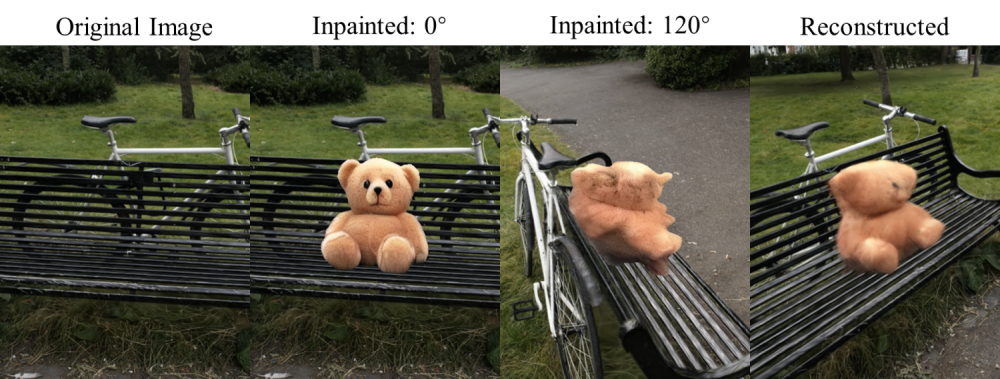}
	\caption{
\textbf{Limitation of 360-degree Content Generation.} Due to the lack of publicly available sufficient training
data, our model encounters challenges in generating and inserting objects across a full 360-degree range. The Wild-RGBD dataset, for instance, provides a maximum of 7,000 usable training samples, with slightly over 100 distinct background scenes available.}
	\label{f_360}
\end{figure}

\begin{figure}
	\centering
	\includegraphics[width=.95\columnwidth]{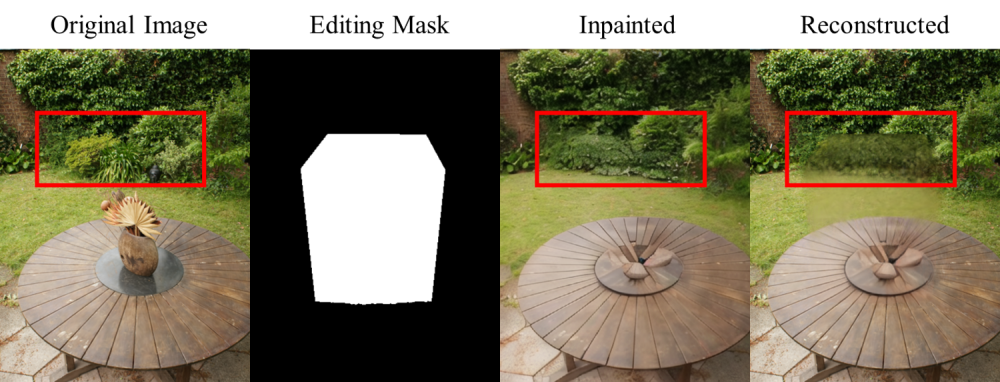}
	\caption{
\textbf{Limitation of Object Removal.}
 Object removal requires patching up the background of the scene rather than creating a new foreground. However, our model cannot ensure the newly-generated background matches precisely with the original background that is already present in the editing region or obscured by the target object, resulting in conflicts during the reconstruction process.}
	\label{f_removal}
\end{figure}

\begin{figure}
	\centering
	\includegraphics[width=.95\columnwidth]{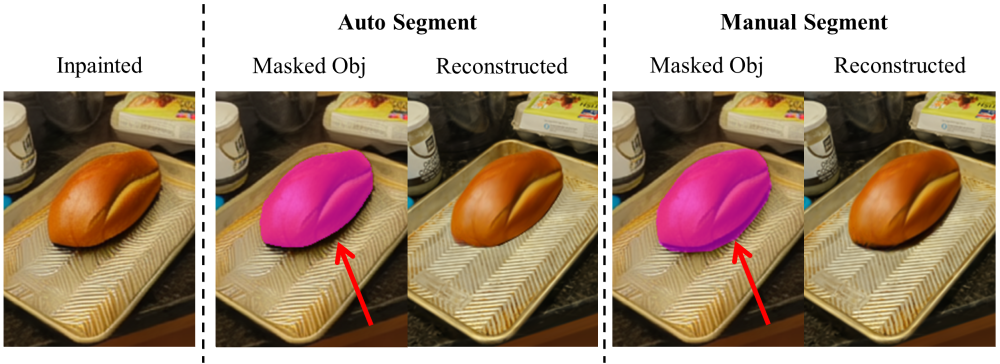}
	\caption{
\textbf{Limitation of Shadow Generation.} 
To preserve the scene background, we segment the target object from the inpainted frames and reintegrate it into the original background for subsequent reconstruction. Nevertheless, the segmentation model may fail to account for shadows, resulting in reconstructions that lack this crucial element.}
	\label{f_shade}
\end{figure}

\section*{CRediT authorship contribution statement}

\textbf{Hongliang Zhong:} Software, Data curation, Formal analysis, Validation, Investigation, Visualization, Methodology, Writing – original draft. \textbf{Can Wang:} Writing – review \& editing. \textbf{Jingbo Zhang:} Writing – review \& editing. \textbf{Jing Liao:} Conceptualization, Resources, Supervision, Funding acquisition, Project administration, Writing – review \& editing.

\section*{Declaration of competing interest}

The authors declare that they have no known competing financial interests or personal relationships that could have appeared to influence the work reported in this paper.

\section*{Declaration of Generative AI and AI-assisted technologies in the writing process}

During the preparation of this work the authors used ChatGPT in order to improve the readability and language of the paper. After using this tool, the authors reviewed and edited the content as needed and take full responsibility for the content of the publication.

\section*{Ethical Approval}

This study does not contain any studies with Human or animal subjects performed by any of the authors.

\section*{Acknowledgments}

The work described in this paper was fully supported by a GRF grant from the Research Grants Council (RGC) of the Hong Kong Special Administrative Region, China [Project No.  CityU 11208123]. 

\bibliographystyle{cas-model2-names}

\bibliography{cas-refs}

\end{document}